# EnviroPiNet: A Physics-Guided AI Model for Predicting Biofilter Performance

Uzma[1*], Fabien Cholet[1], Domenic Quinn[1], Cindy Smith[1], Siming You[1] and William Sloan[1]

[1] James Watt School of Engineering, University of Glasgow, Glasgow, G12 8QQ, UK
Corresponding author email: uzma.k.khan@glasgow.ac.uk

## Abstract

Environmental biotechnologies, such as drinking water biofilters, rely on complex interactions between microbial communities and their surrounding physical-chemical environments. Predicting the performance of these systems is challenging due to high-dimensional, sparse datasets that lack diversity and fail to fully capture system behaviour. Accurate predictive models require innovative, science-guided approaches. In this study, we present the first application of Buckingham Pi theory to modelling biofilter performance. This dimensionality reduction technique identifies meaningful, dimensionless variables that enhance predictive accuracy and improve model interpretability. Using these variables, we developed the Environmental Buckingham Pi Neural Network (EnviroPiNet), a physics-guided model benchmarked against traditional data-driven methods, including Principal Component Analysis (PCA) and autoencoder neural networks. Our findings demonstrate that the EnviroPiNet model achieves an $R^2$ value of 0.9236 on the testing dataset, significantly outperforming PCA and autoencoder methods. The Buckingham Pi variables also provide insights into the physical and chemical relationships governing biofilter behaviour, with implications for system design and optimization. This study highlights the potential of combining physical principles with AI approaches to model complex environmental systems characterized by sparse, high-dimensional datasets.

# Introduction

The success of artificial intelligence (AI) and machine learning (ML) methods depends heavily on computational power and the availability of diverse datasets [1]. With sufficient data and computational resources, AI can uncover patterns and anomalies, often imperceptible to traditional methods [2]. A long-standing goal in engineering is the creation of AI algorithms capable of accurately predicting the behaviour of unbuilt infrastructure or devices. This goal aligns with the foundational aim of engineering theory: to enable accurate prediction [3].

Before the advent of AI, predictive engineering relied on physics-based models, developed through ingenuity and experimentation [14]. These models have successfully informed the design of innovative structures and devices, from skyscrapers to spacecraft and energy-efficient household appliances [5]. However, environmental engineering has struggled to deliver predictive models with comparable reliability. For instance, when wastewater treatment technologies are implemented in new locations or applied to novel waste streams, theoretical models often fail to provide sufficient confidence for full-scale deployment. Instead, extensive pilot studies are conducted to optimize processes empirically before scaling up [6].

This disparity arises partly because environmental systems involve complex interactions between biological and environmental factors, which are difficult to model accurately.

The biological processes at the heart of these systems are influenced by numerous environmental variables, many of which remain poorly understood [7]. Consequently, the development of reliable, physics-based predictive models for environmental technologies has been slow.

Despite these challenges, advances in data collection methods, such as real-time monitoring and molecular microbiology, have made it possible to gather vast amounts of data on environmental systems [8]. These datasets often include high-resolution measurements of physical, chemical, and biological variables [9]. However, a lack of variability in experimental conditions often limits the diversity of these datasets, restricting their utility for fully data-driven models [10]. Sparse and non-exhaustive datasets make it challenging to develop predictive models that generalize well to unseen data [11].

Dimensionality reduction techniques are commonly used to address these challenges by simplifying datasets while preserving essential information [12]. Methods like Principal Component Analysis (PCA) create orthogonal combinations of original variables, capturing most of the variance in a reduced set of components [13]. Similarly, modern techniques like autoencoders use neural networks to encode nonlinear combinations of variables into a reduced feature set, ensuring that the essential characteristics of the data are retained [14]. These reduced features are then used to train predictive models, such as neural networks, with the aim of achieving robust performance on validation datasets. However, such methods remain entirely data-driven and may struggle with sparse or highly variable datasets.

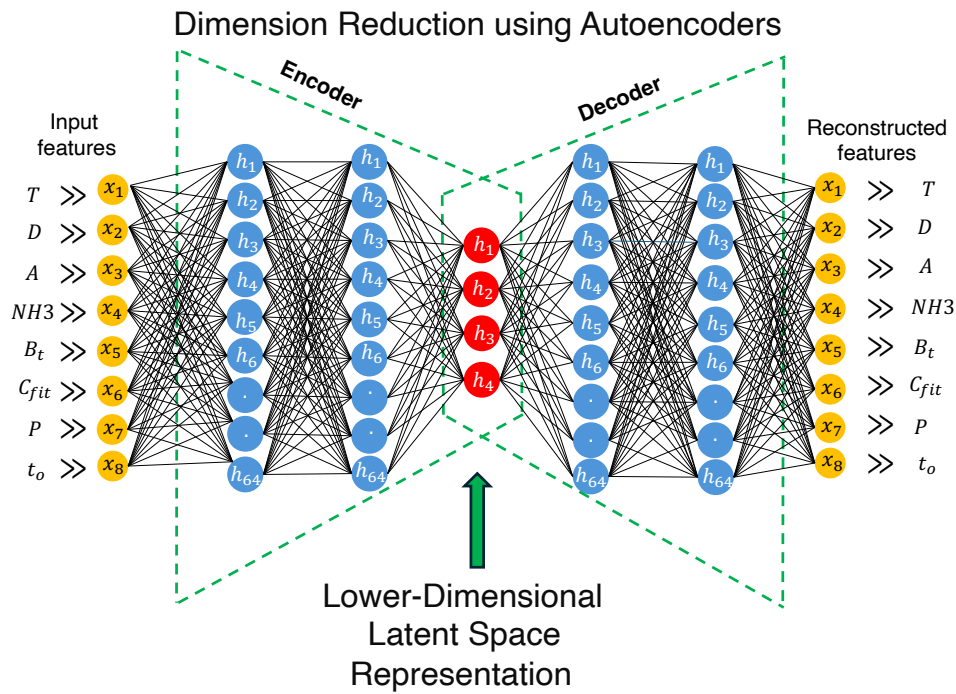

*Fig 1: Autoencoder-based Dimension Reductions. The Autoencoder model reduces the dimensionality of the input data from 8 variables to 4 latent variables. The input variables (Table 1) are temperature, depth, age, ammonia concentration, empty bed contact time, particle size, filter diameter, and balance temperature measured in the hope of inferring the organic carbon in the effluent from biofilters. The decoder ensures that the latent variables can be used to reconstruct the raw data. The latent variables can then be used as inputs to model this and other similar systems.*

In the early 20$^{th}$ century, around the same time as PCA was first implemented, Engineers began to use a dimensional reduction technique, Buckingham Pi theory [15], to make complex multivariate experimental data more tractable. Rather than being purely data driven this method relies on a modicum of understanding of the physics of the system and dimensional analysis, which merely dictates that the units of a formula must balance. If *n* variables in a system are measured and they are quantified using m dimensions (e.g. metres, seconds, kilograms,) then Buckingham Pi theory says that the system can be equally well described by *n-m* non-dimensional variables that are a combination of the originals. It turns out that this seemly innocuous finding is very powerful. It transpired that the set of non-dimensional variables aids our understanding of the complex relationship between variables in physical systems. So, for example, in fluid dynamics the Reynolds number, *Re*, often appears as one the non-dimensional variables. *Re* indicates whether flows are laminar or turbulent and what the degree of turbulence is. Even for flow in channels that are orders of magnitude different in scale or shape, the same value of *Re* implies the same flow regime. Thus, Buckingham Pi theory is routinely used in designing scaled-down experiments of large-scale phenomena. In addition, for certain fluids and aeronautics problems if one assumes a mathematical model for a dependent variable that is the product of Buckingham Pi variables raised to an arbitrary power (monomial) it is possible to fit surprisingly effective models to experimental data. However, it has rarely been used in biological systems, and we have not seen it used as dimensional reduction method in combination with AI for modelling biotechnologies.

Buckingham Pi variables condense information in a reduced set of variables that, by dint of having no dimensions, are a function of properties related by the physics of the system. It could be argued that this is a more intuitive, and potentially, powerful way of reducing the dimensions of a data set than PCA or through the neural network bottlenecks used by autoencoders. In addition, by working with a reduced set of nondimensional variables it has been shown that models can make predictions using variable values that lie out with the range used in training [14]. This means that experimental results made at small-scale in the laboratory might be used to predict a similar system at a much larger scale used in a real application.

In this study, we develop models to predict the concentration of organic carbon in the effluent from drinking water biofilters. We have elected to work with the sort of spare data that are common for biological experiments, where the range and combinations of engineering and biological conditions under which the filters were tested are a small subset of the possible conditions. The training data are from laboratory-scale biofilters, and the test data are from a full-scale operational biofilter. Our goal is to develop a predictive model for biofilter performance by comparing the efficacy of three dimensionality reduction methods followed by different modeling approaches to predict the carbon concentration in the effluent of the filters. We compare: PCA followed by ANN; PCA followed by a linear sum of Buckingham Pi followed by ANN; the engineering approach of deriving Buckingham PI variables and fitting a monomial function of them to the test data; using an autoencoder to derive reduced features and develop a model.

# Methods

# Descriptions of the dataset

To evaluate the models' predictions of organic carbon concentrations in the effluent, three time series datasets were used. The time series are sparse; therefore, we combined all three datasets for training and set aside 20% of the combined data, derived from the third dataset, for evaluation. We based the models on measured variables that were common to all studies and, therefore, potentially useful information in each of the studies had to be discarded.

**Training Dataset:**

Quinn [16] monitored the chemical composition, microbiology, and temperature in the influent effluent and at various depths in a suite of laboratory biofilters. The biofilters were operated for a period of 162 days. Measurements were taken weekly for the first 12 weeks, followed by biweekly measurements thereafter. The filter material was granular activated carbon with almost uniformly dimensioned spherical particles. Ultimately the dataset comprised 175 observations of a vector of variables: $T$, the temperature of the water at the sample location; $Pv$, the pore value of the sample, representing the porosity or void space in the biofilter material; $A$, the age of the filter measured as the time from the start of the experiment; $IC_{org}$, the influent carbon concentration of the sample; $EC_{org}$, the effluent carbon concentration of the sample; $B_t$, the empty bed contact time, which is estimated at the start of the experiment and assumed to be constant for each filter; $P$, is the average diameter of the GAC particles; $C_{fit}$, is the diameter of the filter bed, which is a constant for each filter; $t_o$, is the ambient air temperature, which is assumed to a constant for all locations in the bed at each time point. The units and dimensions of these parameters are shown in Table 1. Each instance is a unique realisation of this vector of variables.

In addition to the Quinn [16] dataset, data from the [17] were incorporated, comprising a total of 26 instances of the same observed variables. The aim of the study was to investigate the effects of biofilter design and operating conditions on the bacterial communities within the filter media and their relationship to the microbiome in the filter effluent water. Specifically, the research focused on how different filter media (granular activated carbon (GAC)-sand versus anthracite-sand) and backwashing strategies (chloraminated versus nonchloraminated water) influenced the bacterial community composition and abundance in pilot-scale biofilters. The study was conducted using pilot-scale biofilter columns, simulating real filter conditions in a controlled laboratory environment. The experimental setup included six replicate columns operated in parallel over an extended period of 18 months. This parallel operation and the reproducibility of sequencing results across the replicate columns enabled a robust assessment of the differences in bacterial communities attributable to the filter media type and operational conditions.

The study from [18], were. also incorporated, comprising 116 samples, of which 54 observations were used for training. This study was conducted on five laboratory-scale biofilters (A, B, C, D and E) packed with GAC (Norit® GAC 1240) which were operated for 48 weeks at 10°C and 20°C degrees in temperature-controlled rooms. Each biofilter had a bed length of 30cm and an internal diameter of 2.6cm and were operated with an EBCT of 3H. All biofilters were fed with the same influent water sourced from a local freshwater reservoir (Pateshill Water Treatment Works, Scottish Water, United Kingdom). Immediately upon collection, the water was filtered through 10 μm polypropylene cartridge filters (Spectrum, UK) and stored in acid-washed and 18.2 M Ohm deionised water rinsed jerry cans at 4∘C until its use for the first 12 weeks, water quality parameter was measured weekly in influent and effluent water. For the remaining 36 weeks, influent and effluent water quality parameters were measured every 6 weeks.

The training dataset combines data from [16,17,18] to enhance its richness and diversity, thereby enabling more robust predictive modelling.

**Testing Dataset**:

To test the models, 20 % of the combined data, derived from the study [18] was set aside. This dataset comprises 62 observations of a vector of variables, which are completely independent of the training dataset. The detail of the test dataset is mentioned in table2. The dimensionless target training data is formed by substituting the dimensional training data into equation (8). The same process is repeated for the dimensionless test data.

Table1: Units and Dimensions of Variables Used in the Training Data.

| Variable | Range | Units | Dimensions |
|---|---|---|---|
| Temperature ($T$) | [4, 28] | k | $L^0.M^0.T^0.K^1$ |
| Pore value (Pv) | [0.1,0.2] | μm | $L^1.M^0.T^0.K^0$ |
| Age ($A$) | [7, 539] | days | $L^0.M^0.T^1.K^0$ |
| Influent Org − CCon ($IC_{org}$) | [3.824, 17] | mg/L | $L^{-3}.M^1.T^0.K^0$ |
| Empty Bed Contact time ($B_t$) | [8.4,450] | Sec | $L^0.M^0.T^1.K^0$ |
| GAC particle diameter ($P$) | [0.9,1.058] | mm | $L^1.M^0.T^0.K^0$ |
| Filter diameter ($C_{fit}$) | [3.75,2.6] | mm | $L^1.M^0.T^0.K^0$ |
| Ambient Temperature ($t_o$) | 37 | k | $L^0.M^0.T^0.K^1$ |
| ECorg ($EC_{org}$) | [0.4,12.729] | mg/L | $L^{-3}.M^1.T^0.K^0$ |

Table2: Units and Dimensions of Variables Used in the Testing Data.

| Variable | Range | Units | Dimensions |
|---|---|---|---|
| Temperature ($T$) | [10, 20] | k | $L^0.M^0.T^0.K^1$ |
| Pore value (Pv) | 0.1 | μm | $L^1.M^0.T^0.K^0$ |
| Age ($A$) | [70, 340] | days | $L^0.M^0.T^1.K^0$ |
| Influent Org − CCon ($IC_{org}$) | [7.725, 21.7] | mg/L | $L^{-3}.M^1.T^0.K^0$ |
| Empty Bed Contact time ($B_t$) | 180 | Sec | $L^0.M^0.T^1.K^0$ |
| GAC particle diameter ($P$) | 1.058 | mm | $L^1.M^0.T^0.K^0$ |
| Filter diameter ($C_{fit}$) | 2.6 | mm | $L^1.M^0.T^0.K^0$ |
| Ambient Temperature ($t_o$) | 37 | k | $L^0.M^0.T^0.K^1$ |
| ECorg ($EC_{org}$) | [5.291, 17.25] | mg/L | $L^{-3}.M^1.T^0.K^0$ |

## Data driven dimensional reduction

The PCA and Autoencoder software used for data driven dimensional reduction are published from algorithms [19] and [20,21], respectively.

**Application of the Buckingham Pi dimensional reduction method and modelling**

For the Buckingham Pi theorem, the algorithm in [22] was implemented. The source code for our implementation is available at https://codeocean.com/capsule/1509576/tree and has been made accessible to reviewers at [23].

Buckingham Pi dimension reduction, whilst well-documented, is not routinely used in biotechnologies or bioinformatics and does involve making subjective choices. In addition, for many engineering applications, the reduced set of dimensionless variables are typically used in a predictive model with a predefined functional form, often a monomial, rather than an AI model. Therefore, the process is outlined by the following steps:

**Initialization**: Postulate that some arbitrary function, *f*, exists that relates the dependent variable of interest, organic carbon, $EC_{org}$, to the vector of independent variables,

$$EC_{org} = f(T, A, Bt, C_{fit}, P, t_o, IC_{org}, Pv) \quad (1)$$

**Define Base Subset**: If the data are described by m dimensions, then select a subset of m independent variables whose units span all m dimensions. In this case there are 4 fundamental dimensions, L, T, M, and K and a base subset of 4 variables with unique units that encapsulate all of the dimensions are T, $IC_{org}$, A and $C_{fit}$ (Table 1).

$$base\ subset = \{T, IC_{org}, A, C_{fit}\} \quad (2)$$

**Dimensional Analysis**: For the remaining m-n independent variables, make the assumption that their dimensions can be expresses as products of powers of the base subset dimensions. In our case, the remaining 5 variables,

$$[Pv] = [T^a IC_{org}{}^b A^c C_{fit}{}^d] \ ,$$
$$[Bt] = [T^e IC_{org}{}^f A^g C_{fit}{}^h] \ , \qquad (3)$$
$$[P] = [T^i IC_{org}{}^h A^k C_{fit}{}^l] \ ,$$
$$[t_o] = [T^m IC_{org}{}^n A^o C_{fit}{}^p] \ ,$$
$$[ECorg] = [T^q IC_{org}{}^r A^s C_{fit}{}^t] \ ,$$

where, $\{a, b, \ldots, t\}$ are arbitrary, as yet, unknown real values. Given equation (3) we can defined a new set of variables,

$$U\pi_1 = \frac{Pv}{T^a\ IC_{org}{}^b A^c C_{fit}{}^d} \ ,$$
$$U\pi_2 = \frac{Bt}{T^e IC_{org}{}^f A^g C_{fit}{}^h} \ ,$$
$$U\pi_3 = \frac{P}{T^i IC_{org}{}^j A^k C_{fit}{}^l} \ , \qquad (4)$$
$$U\pi_4 = \frac{t_o}{T^m IC_{org}{}^n A^o C_{fit}{}^p} \ ,$$
$$Y_\pi = \frac{ECorg}{T^q IC_{org}{}^r A^s C_{fit}{}^t} \ .$$

By insisting the dimensional equivalence in equations (3) we then need to select exponents $\{a, b, \ldots, t\}$ that ensure the new variables are dimensionless.

This becomes an exercise in linear algebra. So, we require

$$[U\pi_1] = \frac{L^1}{(K)^a.(L^{-3}.M^1)^b.(T)^c.L^d} = \left(L^{1+3b-d}.M^{-b}.K^{-a}.T^{-c}\right) = 1$$

$$[U\pi_2] = \frac{T^1}{(K)^e.(L^{-3}.M^1)^f.(T)^g.L^h} = \left(L^{3f-h}.M^{-f}.K^{-e}.T^{1-g}\right) = 1$$

$$[U\pi_3] = \frac{L^1}{(K)^i.(L^{-3}.M^1)^j.(T)^k.L^l} = \left(L^{1-3j-1}.M^{-j}.K^{-i}.T^{-k}\right) = 1 \qquad (5)$$

$$[U\pi_4] = \frac{(K^1)}{(K)^m (L^{-3}.M^1)^n (T)^o (L)^p]} = \left(L^{3n-p}.M^{-n}.K^{1-m}.T^{-o}\right) = 1$$

$$[Y_\pi] = \frac{L^{-3}.M^1}{(K)^q.(L^{-3}.M^1)^r.(T)^s.L^t} = \left(L^{-3+3r-t}.M^{1-r}.K^{-q}.T^{-s}\right) = 1$$

Thus, we equate the exponents. So, for example, for the first Pi group, this process results in equations such as 1+3b-d=0, -b=0, -a=0, and -c=0, or in matrix format,

$$\begin{pmatrix} 0 & 3 & 0 & -1 \\ 0 & -1 & 0 & 0 \\ -1 & 0 & 0 & 0 \\ 0 & 0 & -1 & 0 \end{pmatrix} \begin{pmatrix} a \\ b \\ c \\ d \end{pmatrix} = \begin{pmatrix} 0 \\ 0 \\ 0 \\ 1 \end{pmatrix} \qquad (6)$$

and solving this system gives,

$$\begin{pmatrix} a \\ b \\ c \\ d \end{pmatrix} = \begin{pmatrix} 0 \\ 0 \\ 0 \\ 1 \end{pmatrix}. \qquad (7)$$

So, determining the exponents for all the Pi groups is merely a problem in linear algebra, which has been incorporated into in the supplementary computer code.

**Final Dimensionless Equations**: The exponents can then substitute into equation (4) to obtain expressions for the dimensionless Pi groups and the dependent variable. The resulting equations represent the Buckingham Pi dimension reduction model, where the dependent variable is expressed as a function of the dimensionless Pi groups.

$$\begin{aligned} U\pi_1 &= [Pv/C_{fit}] \\ U\pi_2 &= [B_t/A] \\ U\pi_3 &= [P/C_{fit}] \\ U\pi_4 &= [t_o/T] \\ Y_\pi &= [EC_{org}/IC_{org}] \end{aligned} \qquad (8)$$

In our case the dimensionless Pi groups emerge as simple ratios of the original variables. Now, $y_\pi$ becomes a function, $g$ say, of $U\pi_1, U\pi_2$, and $U\pi_3$ expressed as

$$Y_\pi = g(U\pi_1, U\pi_2, U\pi_3, U\pi_4) \qquad (9)$$

The function g is our predictive model. The, N, observations in the training dataset $[U\pi_1(i), U\pi_2(i), U\pi_3(i), U\pi_4(i), Y_\pi(i)]_{i=1}^N$ can then be used to calibrate a function or train an algorithm to derive an appropriate function, $g$.

**Determine the model (function, g)**: Having reduced the dimensions it remains to determine the function g. For this we can use a wide range of approaches. We can impose a particular functional form, which is the approach traditionally used in physical engineering applications of Buckingham Pi theory; typically, a monomial function is the first to be applied. Alternatively, we can use entirely data driven models, where the ultimate model structure is more or less apparent depending on the methods uses.

So, for the monomial approach we assume that,

$$Y_\pi = \alpha_0 \; U\pi_1^{\alpha_1} U\pi_2^{\alpha_2} U\pi_3^{\alpha_3} U\pi_4^{\alpha_4} \; , \qquad (10)$$

where, $\{\alpha_1, ..., \alpha_4\}$ are arbitrary constants. To estimate these constants based on the training data then we note,

$$\ln(Y_\pi) = \ln(\alpha_0) + \sum_{i=1}^{4}(\alpha_i \ln(U\pi_i)) , \qquad (11)$$

and, therefore, we can simply use linear regression. This transformation allows us to use linear regression to estimate the values of $\{\alpha_1, ..., \alpha_4\}$ where each $\alpha_1$ is treated as a coefficient to be learned from the data. However, to mitigate overfitting and prevent the model from excessively fitting the training data, ridge regression [24] was applied to regularize the model. Ridge regression applies a penalty to the size of the coefficients, helping to shrink them and improve generalizability.

To determine the optimal regularization strength, $\alpha_1$, a grid search [25] with cross-validation was performed over a range of values for the penalty parameter. The best $\boldsymbol{\alpha_1}$ was selected based on the performance of the model in terms of minimizing **the** mean squared error during the cross-validation process. By applying ridge regression with the optimal value of $\alpha_1$, the model's coefficients were regularized, leading to more stable and robust estimates for the constants $\{\alpha_1, ..., \alpha_4\}$. We use BP-LR to refer to this predictive model. The purely data driven method that we have adopted is to use a neural network. Using the same reduced non-dimensional training set we first scale the non-dimensional variables to have a mean of zero and a standard deviation of one. The scaled independent variable vectors then serve as inputs for a three-layer feedforward neural network [26]. This network comprises one input layer, two hidden layers, and one output layer, with ReLU activation functions used for all layers except the output layer, which employs a linear activation function. The architecture of the neural network is defined by the parameters listed in Table 3, where the number of seeds is set to 7. This "number of seeds" controls randomization aspects like weight initialization and data shuffling, ensuring consistent results across training runs for reproducibility. Additionally, the Adaptive Moment Estimation (Adam) optimizer is used for training the model. Note that we use a 5-fold cross validation approach to evaluate the model's performance. In this process, the training dataset is divided into five subsets, and each subset is used once as a validation set while the remaining four subsets are used for training. This fitting method combines training with validation each time the training data set is presented to the network (epoch), ensurining reliable performance evaluation and preventing overfitting. Additionally, an L2 kernel regularizer is applied to the network to mitigate overfitting and enhance generalization performance. We refer to this application of the neural networks to Environmental Buckingham Pi variables as EnviroPiNet.

Table 3: Training hyperparameters

| Parameter | Value |
|---|---|
| Batch size for training | 16 |
| Learning rate for optimization | 0.0001 |
| Number of units in each layer | 64 |
| Total number of layers (1 input, 2 hidden) | 3 |
| Number of training epochs | 50 |
| Number of splits for cross-validation | 5 |
| Activation function | ReLU |
| L2 regularization parameter | 0.01 |
| Number of random seeds for reproducibility | 7 |

**Determining the models for the PCA and Autoencoder reduced dimensions**

The Buckingham Pi dimensional reduction method delivered 4 dimensionless variables from the original 8 independent variables. Thus, we applied PCA and retained the first 4 principal components {$PCA_1,...PCA_4$} as a reduced set of variates. Similarly for the autoencoder the bottleneck (Fig. 1) comprised 4 features {$h_1,....,h_4$} that constituted a reduced set of variates. The challenge then, as with the Pi variates, was to determine arbitrary functions, $p$ and $q$ say, such that relate out independent variable, $EC_{org}$, to the reduced variates.

$$EC_{org} = p(PCA_1, PCA_2, PCA_3, PCA_4) , \qquad (11)$$
$$EC_{org} = q(h_1, h_2, h_3, h_4) . \qquad (13)$$

In both cases for , p and q, we used a neural network constructed in exactly the same way as for the Buckingham Pi analysis. These models are refered to as PC-NN and Autoencoder-NN. For conmpleteness we also use monomial mode for p and q in equations 12 and 13, which we refer to as PC-LR and Autoencoder-LR respectively.

## Measures of performance

The proposed model's evaluation employs both sMAPE and $R^2$ metrics [27] to provide comprehensive insights into its performance. By considering multiple evaluation metrics, we can better assess the model's accuracy and generalization capability across different datasets and scenarios.

### R-squared ($R^2$)

During training, the model learns the relationships between the input variables and the target variable by adjusting its parameters to minimize the prediction error, typically measured using the Mean Squared Error (MSE) [28]. The R-squared ($R^2$) metric is utilized to evaluate the performance of the designed predictive model.

$$R^2 = 1 - \frac{\sum_{i=1}^{n}(y_i - \hat{y}_i)^2}{\sum_{i=1}^{n}(y_i - \bar{y})^2} \qquad (14)$$

where $y_i$ are the observed values, $\hat{y}_i$ are the predicted values and $\bar{y}$ is the mean of the observations. $R^2$, also known as the coefficient of determination, quantifies the proportion of the variance in the dependent variable that is predictable from the independent variables. Its value ranges from 0 to 1, with a value of 1 indicating a perfect fit, where the model explains all the variability in the dependent variable. A value of 0 suggests that the model does not explain any of the variability, while negative values indicate that the model fits the data worse than a horizontal line. $R^2$'s bounded nature facilitates consistent interpretation across different datasets and applications, making it particularly valuable when comparing predicted and measured values of continuous variables.

### Symmetric Mean Absolute Percentage Error (sMAPE)

In addition to R-squared, the Symmetric Mean Absolute Percentage Error (sMAPE) is employed as a metric to evaluate the model's performance. sMAPE offers advantages such as simplicity and computational efficiency.

sMAPE is calculated as:

$$\text{sMAPE} = \frac{100}{n} \sum_{i=1}^{n} \frac{|y_i - Y_i|}{|y_i| + |Y_i|} \quad , \qquad (15)$$

where $y_t$ represents the true value and $Y_t$ represents the predicted value, and $n$ is the number of samples. However, sMAPE values may vary significantly from one dataset to another, making direct comparisons challenging.

# Results

In this section, we delve into the outcomes of applying our various reduced dimension models to forecast the concentration of organic carbon in drinking water biofilter effluent. The first stage is to derive the reduced set of four variants using Bukingham Pi (equation 8), Principal Components (equation 10) and Autoencoder (equation 11). Generic code for achieving this using any dataset has been depostited in Github https://github.com/uzmauzma/Biofilter-Performance-ML-Buckingham-Pi. These are then used as inputs to the Neural Network or to the linear regression that is used to determine the coeffiecients of the monomial model. The training of the neural network used 5-fold cross validation so that after every epoch a measure of goodness of fit can be calculates for a randomly allocated 80% portion of the data that is used for training and a second measure can be calculated the remaining 20% of the data. The graph of these two measures is called the loss plot. Fig. 2 shows loss plots for EnviroPiNet, using 2 measures: Mean Squared Error (MSE) (Fig. 2A) and the Mean Absolute Error (MAE) (Fig. 2B). These curves illustrate the convergence of both training and validation measures, suggesting that the model effectively learns from the training data and generalizes well to unseen validation data. For the PCA-NN and Autoencoder-NN models, the loss curves in Fig.3 show that the model learns effectively from the training data and generalizes well on the validation data. However, as seen in Table 3, it does not perform well on unseen test data. This could be because, during training, the model is trained, tuned, or adjusted based on the validation performance. Since PCA and autoencoders focus on capturing the most significant variance in the data, they may ignore less prominent but still important features that are crucial for predicting the target variables. Therefore, the validation data may exhibit the same general patterns as the training data, which allows the model to perform well during training. However, when we apply the model to completely unseen test data, the test data may contain patterns or variations that were not captured well by the dimensionality reduction techniques, leading to poor performance on the unseen data.

Typically K-fold cross validations is not applied for regression analyses, rather a single goodness of fit is reported for the model using training data.

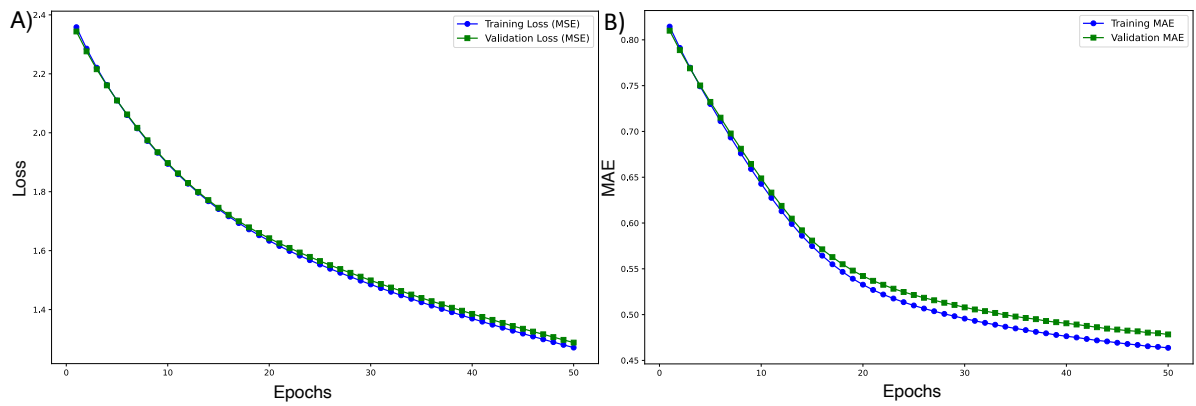

**Fig 2**: (A) Learning curve depicting the loss by Mean Square Error (MSE) on **EnviroPiNet**. The curve illustrates a decrease in both training and validation loss, with both curves converging. This convergence suggests that the model is effectively learning from the training data and generalizing well to unseen validation data. (B) Model performance depicted by Mean Absolute Error (MAE) on **EnviroPiNet**.

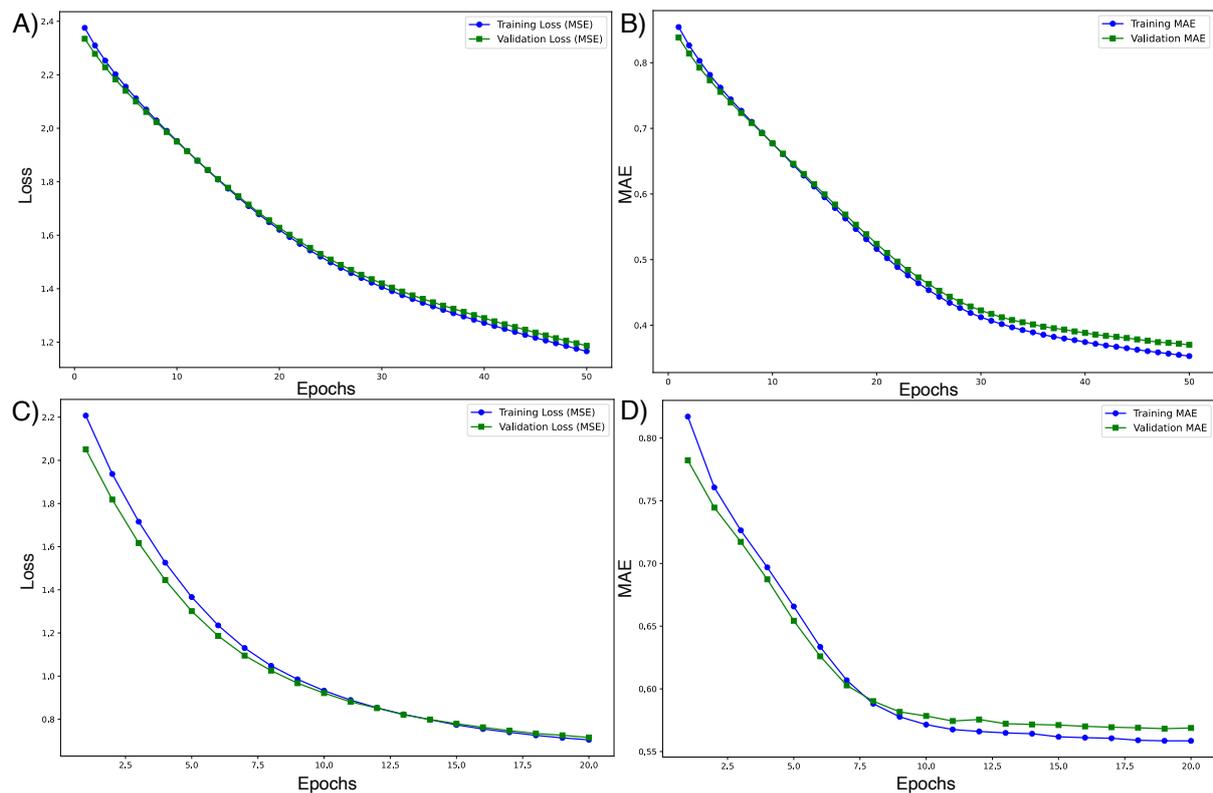

**Fig 3**: (A) and (C) Learning curves showing the Mean Squared Error (MSE) for Autoencoder and PCA models. Both training and validation losses decrease steadily and converge, indicating effective learning and good generalization to the validation data. (C) and (D) Model performance shown by Mean Absolute Error (MAE) for Autoencoder and PCA, with lower MAE values reflecting better performance on the validation data.

Having trained the models they are applied to the unseen test dataset, which are the data from the [18]. Thus, the reduced variates are calculated from the raw data and then the trained neural neworks, with trained node weights unchanged, and the monomial models, with the trained exponents, are applied. Again, we can calculate the various performance metrics can be calculated. The R² and sMAPE metrics are compared for the models appled to the trainind and test datasets in Tables 3 and 4 respectively.

Table 3: **R-squared ($R^2$)** is utilized to assess the predictive performance of the models for both training and evaluation datasets, with the results averaged across seven different seeds.

| Comparisons algorithm | Training | Testing |
|---|---|---|
| PCA-NN | 0.76 | -1.57 |
| PCA-LR | -0.03 | -2.30 |
| Autoencoder-NN | 0.47 | 0.35 |
| Autoencoder-LR | -0.07 | -2.22 |
| EnviroPiNet | 0.95 | 0.92 |
| BP-LR | 0.4 | 0.34 |

Table 4: **sMAPE** employed for predicting performance on both training and evaluation datasets, with results averaged across seven different seeds.

| Comparisons algorithm | Training | Testing |
|---|---|---|
| PCA-NN | 12.8 | 26.8 |
| PCA-LR | 25.8 | 33.7 |
| Autoencoder-NN | 24.0 | 18.5 |
| Autoencoder-LR | 27.0 | 32.5 |
| EnviroPiNet | 2.31 | 4.5 |
| BP-LR | 18.2 | 10.8 |

For Autoencoder reduced set of variates the relationship to the original variables can be nonlinear and is difficult to interpret in any biological or physical way. The reduced set of PCAs are orthogonal linear combinations of the orginal expanatory variables and thus are slightly easier to interpret in terms of the biology and physics. Buckingham Pi variates are, in effect, selected to encapsulate the science of the system. Thus, especially when used in an analytic functional form, their influence on the independent variable can be interpreted. The linear regression [28] fitted exponents for the BP-LR are reported:

$$\frac{EC_{org}}{IC_{org}} = \text{intercept} \left(\frac{Pv}{C_{fit}}\right)^{-0.06} \left(\frac{B_t}{A}\right)^{-0.30} \left(\frac{P}{C_{fit}}\right)^{0.53} \left(\frac{t_0}{T}\right)^{-0.09} \quad (16)$$

where the intercept is the estimated constant, given by exp (-0.07) = 0.93. and each term in the equation represents the effect of different variables on the organic carbon concentration in the treated water.

The model suggests that changes in $(\frac{P}{C_{fit}})$ have a large effect on the proportion of carbon removed in the filter. Thus, the smaller radius of GAC participles and the greater the radius of the filter (a surrogate for the amount of GAC) the more carbon removed. Given that this variable related to the surface are of activated carbon, this make sense, The ratio $(\frac{B_t}{A})$ is raised to the power -0.3 which means that as A increases the proportion of carbon removed decreases and so, the older the filter the poorer the treatment, $Pv$ and Temperature changes have a small effect on carbon removal.

In addition to the results presented, Fig.4(A) and 4(B) provide visual representations of the linear regression plots for the test dataset, comparaing the EnviroPiNet and BP_LR models. Fig.4(A) showcases the Pearson correlation coefficient (r-value) of 0.97 with $p < 0.05$ for EnviroPiNet, along with an R-squared value of 0.92. Fig. 4(B) demonstrates an r-value of 0.86 with $p < 0.05$ for BP_LR along with an R-squared value of 0.34. These figures emphasize the distinct characteristics and performance metrics of each model when evaluated on the test data.

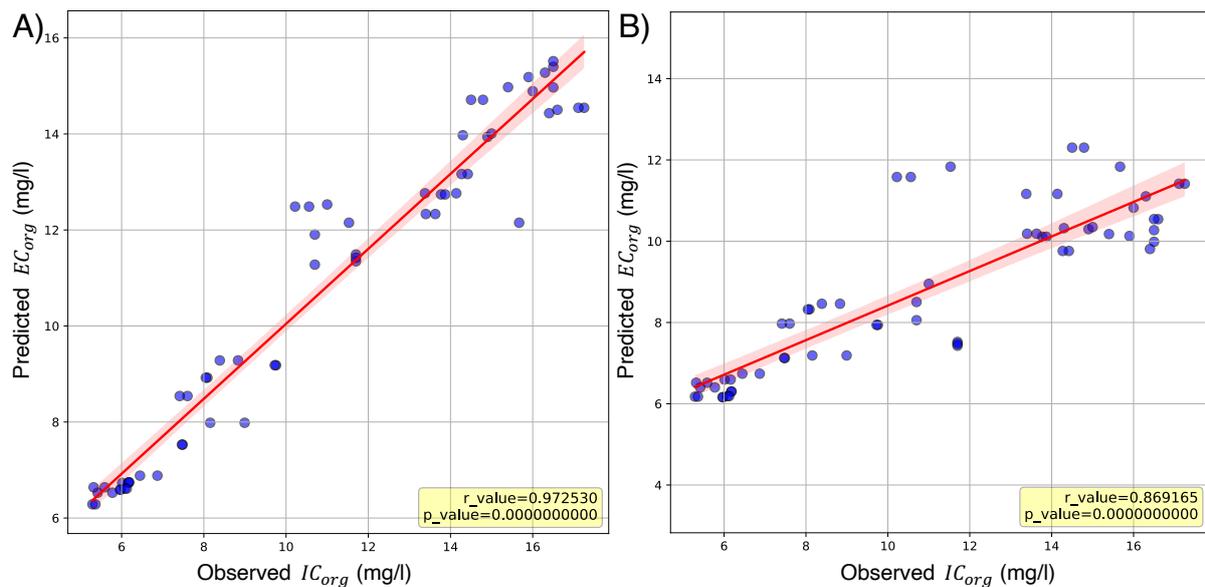

**Fig 4(A):** The regression plot generated using EnviroPiNet depicts a Pearson correlation coefficient (r-value) of 0.97 with $p < 0.05$. Conversely, Fig 4(B) illustrates the regression plot generated using BP_LR, showing an r-value of 0.86 with $p < 0.05$.

In this study, we aimed to predict the carbon concentration $EC_{org}$, in drinking water biofilter effluent. The Environmental Buckingham Pi Neural Network (EnviroPiNet) demonstrated superior performance for predicting $EC_{org}$, achieving an R² value of 0.92, as shown in Table 3, and a sMAPE value of 4.5, as presented in Table 4. The model also yielded a Pearson correlation coefficient r -value) of 0.97 with a p-value less than 0.05, as depicted in Fig. 4(A). The corresponding learning curve is illustrated in Fig. 2(A).

In comparison, linear regression (BP-LR) was applied, resulting in a Pearson correlation coefficient r-value of 0.86 with a p-value less than 0.05, as shown in Fig.4(B), an R² value of 0.34, as presented in Table 2, and a sMAPE value of 10.8, as shown in Table 3, for the prediction of $EC_{org}$,

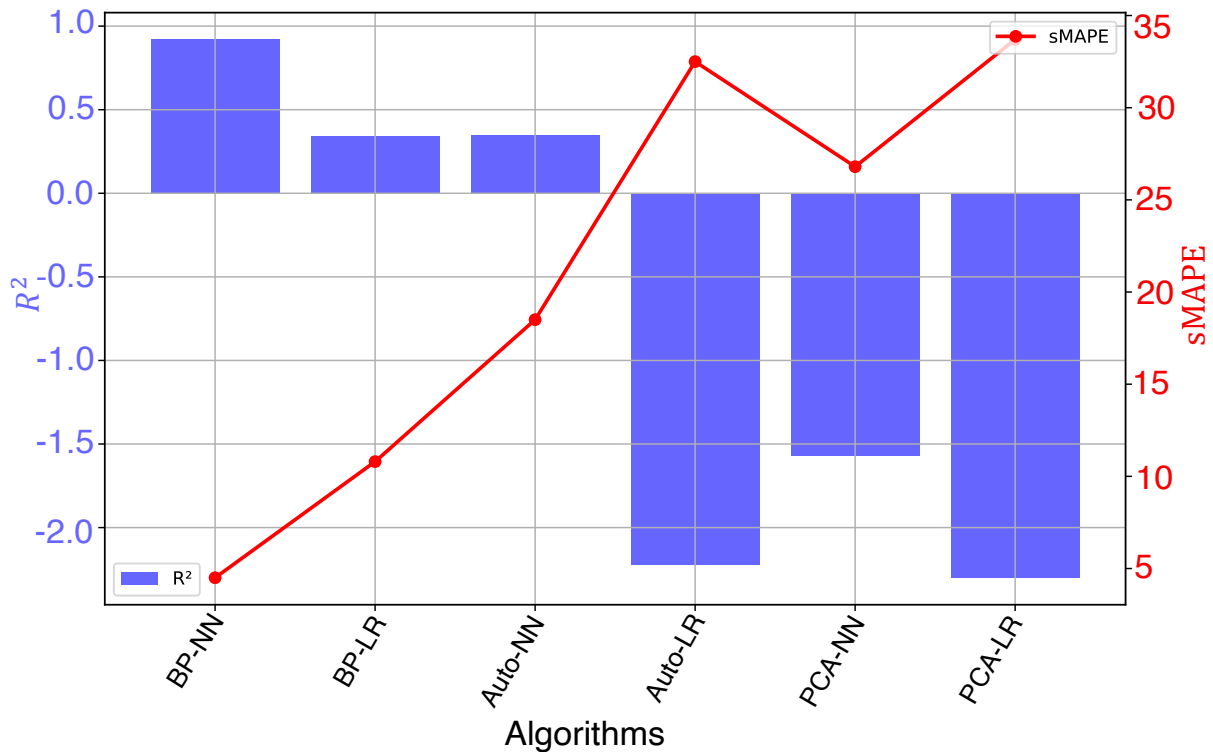

**Fig 5**: Comparison of algorithms on the testing dataset using two performance metrics: $R^2$ and sMAPE. The $R^2$ metric shows higher values for EnviroPiNet and BP-LR, indicating better performance, while the sMAPE metric highlights both EnviroPiNet and BP-LR with the lowest values. Overall, EnviroPiNet outperforms BP-LR across both metrics, demonstrating superior performance.

The results show that the EnviroPiNet model's $R^2$ value outperformed the linear regression model. This suggests that the EnviroPiNet benefited from the Buckingham Pi theorem, enabling it to leverage the advantages of learning from a low-diversity dataset while using the neural network to capture nonlinear and complex relationships. Consequently, it outperformed linear regression in this specific application, as shown in Fig.5.

In discussing our findings, it's essential to underscore the significance of our comparative analysis across various dimensionality reduction techniques and modeling approaches. By systematically evaluating methods such as BP, PCA, and autoencoder, in conjunction with neural networks or a monomial function of the reduced variates, we gain valuable insights into their respective strengths and weaknesses in predicting carbon concentrations in drinking water biofilter effluent.

PCA and autoencoder methods are generally designed to reduce dimensionally without necessarily considering the physical or interpretive meaning of the features. As a result, they might produce future combinations that aren't as relevent for predicting corbon concentrations in water, leading to suboptimal generalization. However, The Buckingham Pi approach maintains physical relevance, which helps the model understand and learn the meaningful relationships in the data. This results in better generalization and performance on test data.

Our assessment comprised two distinct analyses aimed at scrutinizing the efficacy of the model. Employing R-squared ($R^2$) as a performance metric (Table 3), we observed promising outcomes when employing the Buckingham Pi theorem for dimensionality reduction followed by neural network application. Specifically, EnviroPiNet achieved an $R^2$ value of 0.92, outperforming PCA (-1.57) and autoencoder (0.35). Even when linear regression was applied post-reduction, BP still demonstrated favorable performance, with an $R^2$ value of 0.34 compared to PCA (-2.30) and autoencoder (-2.22). These findings underscore the superior dimensionality reduction capabilities of BP in predicting carbon concentrations in effluent water, especially when combined with neural network techniques.

Analysis using sMAPE as a performance metric (Table 4) demonstrated that Buckingham Pi, followed by ANN (EnviroPiNet), exhibited the best performance, achieving an sMAPE value of 4.5. Comparatively, PCA and autoencoder achieved higher sMAPE values of 26.8 and 18.5, respectively. Similarly, Buckingham Pi followed by linear regression (BP-LR) also performed better, achieving an sMAPE value of 10.8, outperforming autoencoder (32.5) and PCA (33.7).

Furthermore, our comparative analysis highlights that while R-squared reflects the overall predictive accuracy of the models, sMAPE provides insights into the accuracy of individual predictions. In both performance metrics, Buckingham Pi (BP) consistently outperformed other methods, demonstrating its strengths in predictive accuracy and individual prediction precision. The study [26] supports the preference for R-squared over sMAPE when comparing predicted and measured values of continuous variables. R-squared offers bounded values that are interpretable and comparable across datasets, making it a more consistent metric for evaluating the overall performance of predictive models, particularly when the focus is on general trends and fit rather than individual prediction accuracy.

Implications extend beyond academia to real-world applications in water quality management. Accurate prediction of organic carbon concentrations serves as an early warning system for potential contaminants in drinking water, enabling proactive measures to safeguard public health and optimize treatment processes.

Importantly, while these dimensionality reduction techniques effectively predict dimensionless variables (Pi variables), post-processing is necessary to revert the predictions into meaningful dimensional variables such as $EC_{org}$. This step highlights the practical implications of these predictions, which indicate that EnviroPiNet showed a higher correlation and a better R² value of 0.92, indicating that Neural Network was able to capture the relationship between the predictors and the target variable more effectively than BP-LR in this case.

Our study has provided valuable insights, but future research should focus on evaluating the model across different test datasets to assess its robustness and generalizability. This will ensure comprehensive evaluations and enhance the applicability of our approach. Since this is the first time the model has been applied in biofilters, future work could explore its application to similar systems or processes to further validate its predictive capabilities and expand its scope. Additionally, refining the model architecture and incorporating advanced techniques could improve its predictive performance.

By addressing these limitations and pursuing avenues for future research, we can continue to advance our understanding of water quality management and contribute to the development of more effective and sustainable treatment strategies.

In conclusion, our comparative analysis underscores the effectiveness of different dimensionality reduction techniques and modeling approaches in predicting carbon concentrations in drinking water. The superior performance of EnviroPiNet holds promise for enhancing water quality management practices, ultimately benefiting public health and environmental sustainability.